\documentclass[runningheads]{llncs}

\usepackage[utf8]{inputenc} 
\usepackage[T1]{fontenc}    
\usepackage{hyperref}       
\usepackage{url}            
\usepackage{booktabs}       
\usepackage{amsfonts}       
\usepackage{nicefrac}       
\usepackage{microtype}      
\usepackage{times}          
\usepackage{helvet}         
\usepackage{courier}        
\usepackage{url}            
\usepackage{graphicx}       
\usepackage{wrapfig}
\usepackage{natbib}
\usepackage{color}
\usepackage{float}
\usepackage{subcaption}
\usepackage{algorithm, algpseudocode}

\begin{document}

\title{Multi-Objective Automatic Machine Learning with AutoxgboostMC}

\author{
    Florian Pfisterer        \and
    Stefan Coors             \and
    Janek Thomas             \and
    Bernd Bischl
}

\authorrunning{Pfisterer et al.}
\institute{LMU Munich \\}

\maketitle

\begin{abstract}
AutoML systems are currently rising in popularity, as they can build powerful models without human oversight. They often combine techniques from many different sub-fields of machine learning in order to find a model or set of models that optimize a user-supplied criterion, such as predictive performance. The ultimate goal of such systems is to reduce the amount of time spent on menial tasks, or tasks that can be solved better by algorithms while leaving decisions that require human intelligence to the end-user. In recent years, the importance of other criteria, such as fairness and interpretability, and many others have become more and more apparent. Current AutoML frameworks either do not allow to optimize such secondary criteria or only do so by limiting the system's choice of models and preprocessing steps. We propose to optimize additional criteria defined by the user directly to guide the search towards an optimal machine learning pipeline. In order to demonstrate the need and usefulness of our approach, we provide a simple multi-criteria AutoML system and showcase an exemplary application.
\end{abstract}

\section{Introduction}

While many stages of a data analysis project still need to be done manually by human data scientists, other parts, such as model selection and algorithm configuration can be efficiently handled by algorithms.
This does not only reduce the time required by humans, but also allows to leverage parallelization.
A typical challenge is the selection of appropriate algorithms and corresponding hyperparameters for a given problem.
Multiple methods for solving this \textit{Combined Algorithm Selection and Hyperparameter optimization (CASH)} problem \citep{autoweka} already exist and are typically referred to as \textit{Automatic Machine Learning} (AutoML).

There is a growing number of approaches for AutoML available to non-specialists.
As one of the first frameworks, Auto-WEKA ~\citep{autoweka} introduced a system for automatically choosing from a broad variety of learning algorithms implemented in the open source software \textit{WEKA} ~\citep{weka}.
Auto-WEKA simultaneously tunes hyperparameters over several learning algorithms using the Bayesian optimization framework SMAC ~\citep{smac}.
Similar to Auto-WEKA is \textit{auto-sklearn} ~\citep{autosklearn}, which is based on the scikit-learn toolkit for python and includes all of its learners as well as available preprocessing operations.
It stacks multiple models to achieve high predictive performance.
Another python-based AutoML tool is called \textit{Tree-based Pipeline Optimization Tool (TPOT)} by ~\citet{tpot} and uses genetic programming instead of Bayesian optimization to tune over a similar space as auto-sklearn.

In this work we consider an approach that configures a \textit{machine learning pipeline}, i.e. an approach that optimizes pre- and post-processing steps along with algorithm hyperparameters of a single gradient boosting model \citep{gradientboosting}.
By focusing on a single learning algorithm, hyperparameters can be optimized much more thoroughly and the resulting model can be analyzed and deployed more easily. 
Gradient boosting models can vary from very simple to highly complex models through the choice of appropriate hyper-parameters. 
Single learner systems reduce the complexity of the configuration space, however, a drawback is, that this search-space possibly does not include optimal configurations, benefits from stacking and ensembling are not explored, and that thus optimal predictive performance may not be achieved.\\
Only few single-learner AutoML methods exist.
The \textit{autoxgboost} software proposed by \citet{autoxgboost} is a single-learner strategy using the \textit{xgboost} (\cite{xgboost}) algorithm, with model based optimization for hyperparameter tuning.
The success of such single-learner strategies was shown in the NIPS 2018 AutoML Challenge ~\citep{guyon2019}: The winning entry, \textit{AutoGBT} (\citet{AutoGBT}) only used LightGBM \citep{lightgbm} models with a simple preprocessing scheme. \\

Several new challenges occur when adapting AutoML systems for multi-criteria optimization. 
Depending on the objective to be optimized, different pre- and post-processing methods might be required in order to obtain optimal performances. 
Additionally, different user-preferences regarding which trade-offs between objectives a user is willing to make have to be incorporated. 
We argue that giving the user the opportunity to intervene in the process can be beneficial here. 
Lastly, measures that quantify objectives such as \textit{fairness}, \textit{interpretability} and \textit{robustness} are often not readily available.
In this work, we want to $i)$ emphasize the necessity for considering multiple objectives in AutoML, $ii)$ provide several measures that can be useful in such a context and $iii)$ propose a simple system that allows the user to automatically optimize over a set of measures.
In contrast to previous work, we focus on optimizing multiple user-defined criteria simultaneously.
Being able to transparently optimize multiple criteria is a crucial missing step in many existing frameworks.
In order to underline the need for several different criteria, we demonstrate the functionality of our proposed framework in a practical use case.

\section{The case for additional criteria in AutoML}

Multi-criteria optimization is well-established in machine learning for example in ROC analysis (\cite{everson2006multi}), computational biology (\cite{handl2007multiobjective}) and other fields. \cite{jin2008pareto} study various use cases, among others, models are optimized jointly with respect to interpretability and predictive performance.
Multi-criteria optimization is also actively researched in the field of Algorithm Configuration. \cite{blot16} introduce a multi-criteria iterative local search procedure for configuring SAT solvers, while \cite{sprint} introduce a racing-based approach.
Different Multi-criteria Bayesian Optimization approaches have also been proposed (c.f. ~\cite{multicritMBO}), but it has not been thoroughly investigated as a part of AutoML frameworks until now.
Many different algorithms, such as  approaches based on iterated local search or racing as well as genetic algorithm based approaches can be used to optimize machine learning pipelines, given that they can deal with hierarchical mixed continuous and discrete spaces.
We choose Bayesian Optimization because it has been shown to work well with relatively small budgets \citep{mlrMBO} and complex hierarchical spaces can be optimized by using random forests as surrogate models.\\

Only being able to optimize a single performance measure entails multiple pitfalls that can possibly be avoided when multiple performance measures are optimized jointly.
This has been emphasized recently in the \textbf{FatML} (Fairness, Accountability, and Transparency in Machine Learning) community which made the case for models that emphasize transparency and fairness \citep{fairmlbook}.
The need for models that do not discriminate against parts of the population in order to achieve optimal predictive performance has garnered widespread support, yet no real options that allow users to jointly search for fair, transparent and well-performing models are available. 
The case for other criteria, that might be relevant to a user has been made in many other areas of machine learning. 
Examples include models that emphasize sparseness, a lower inference time, i.e., when searching for items in databases ~\citep{FAISS}, a low memory footprint, for example when deploying models on mobile phones ~\citep{MobileNet} or a combination of those when doing inference on edge devices (\cite{SpeedAccObjDetect}).
Similarly, the case for requiring \textit{robust} models, i.e., models that are robust to \textit{perturbations} in the data (adversarial perturbations, c.f ~\cite{Papernot15}) can be made.
Models that satisfy a user-desired trade-off might not be found using single-criteria optimization (\cite{jin2008pareto}). 
It is important to distinguish between jointly optimizing multiple optimization criteria, 
and constrained optimization (c.f. \cite{hernandez2016general} for an overview).
Achieving a certain model size might be paramount to be able to deploy a machine learning pipeline to a end user device, but having a model smaller than this size threshold is only of minor interest. 
The concept of constraints in multi-criteria optimization is a well researched topic (c.f. \cite{fan2017comparative}). 

\subsection{Human in the loop approaches in AutoML}

The original aim of AutoML systems is to transfer the CASH problem from the hands of a human to the machine. This does not only allow experienced machine learning researchers to focus on other tasks, such as validating data and feature engineering leveraging domain knowledge, but also enables a broader public to apply Machine Learning, as steps that require a machine learning expert, like selecting algorithms and tuning their hyperparameters, are fully automated. 
The AutoML system is thus treated as a black-box, that can only be influenced by some hyperparameters at the beginning of the training, essentially removing the human from the optimization loop.

In situations, where multiple criteria have to be optimized simultaneously, a trade-off between the different measures is often required. 
Specifying this trade-off \textit{a priori} can be difficult when possible trade-offs are not known. 
\cite{Hakanen2017OnUD} propose an interactive Bayesian Optimization extension to parEgo \citep{parego}, that allows a user to iteratively select preferred ranges for the different optimization criteria. This does not only allow to search for solutions in the region a user is interested in, but also allows the user to adapt preferences throughout the procedure.
This emphasizes the need for users to guide the AutoML process, essentially putting the user back into the loop, albeit in a different fashion. Instead of manually configuring the pipeline, the user is now able to occasionally supervise the search process and make adjustments where needed. A different approach, that allows the user to guide the search process by adapting the search space and tries to visualize and explain decisions made within AutoML systems has been proposed in ~\citep{ATMSeer}.\\

\subsection{Measures for Multi-Criteria AutoML}
\label{sec:measures}
\begin{figure}[ht]
\begin{minipage}{0.6\textwidth}
\centering
\includegraphics[width=0.95\textwidth, page = 3]{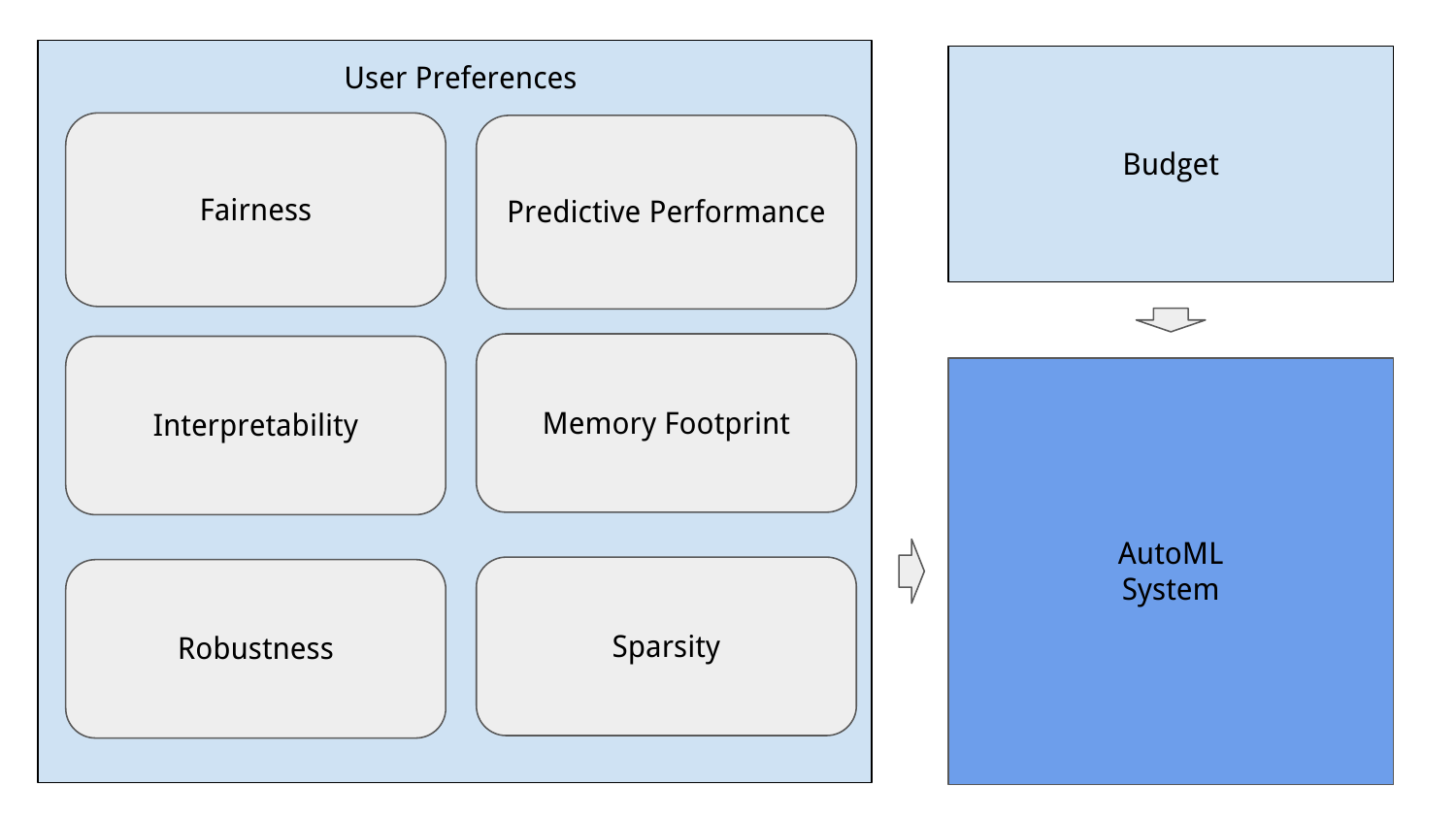}
\caption{User Input to AutoML Systems.}
\centering
\label{fig:workflow}
\end{minipage}
\begin{minipage}{0.4\textwidth}
Multi-criteria optimization methods usually explore the whole pareto front defined by trade-offs between the different objectives.
In our work, we mainly allow the user to guide the search process in two ways:  We enable the user to focus on exploring different parts of the pareto front by selecting upper and lower trade-offs relevant to the user. Second, we allow the user to adapt the search space, by adjusting hyperparameter ranges and activating or deactivating processing steps. This allows the user to shape the result towards personal preferences.
\end{minipage}
\end{figure}
In order to start the investigation into Multi-Criteria AutoML, we aim to provide a list of measures that cover a wide variety of use-cases. 
We want to stress, that the proposed measures are not comprehensive or final, but instead can be thought of as exchangeable building blocks that can serve as a useful proxy in the AutoML process.
We hope to emphasize the necessity for measures, that better reflect the underlying model characteristics we aim to optimize.

\paragraph{Predictive Performance} can be quantified using many different measures, such as Accuracy, F-Score or Area under the Curve for classification and Mean Squared Error or Mean Absolute Error for regression. As those measures are already widely known, we refrain from going into more detail in this work.

\paragraph{Interpretability}
In order to  make a machine learning model's decisions more transparent, different methods that aim at providing human-understandable explanations have been proposed (c.f. ~\cite{molnar2019}). 
Many of those work in a \textit{model-agnostic} and \textit{post-hoc} fashion, which is desirable for AutoML processes, as this allows the user to explain arbitrary models resulting from AutoML processes. Interpretability methods can produce misleading results if a model is too complex. Quantifying interpretability, i.e. determining how complex predictive decisions of a given model are could be a first step, making it a useful criterion to optimize for AutoML systems. A first approach has been proposed in ~\cite{molnar2019quantifying}, describing $3$ measures that can be used as a proxy for interpretability. We implement those measures and briefly present each:

\begin{itemize} 
 \item \textbf{Complexity of main effects} ~\cite{molnar2019quantifying} propose to determine the average shape complexity of ALE ~\citep{apley2016visualizing} main effects by the number of parameters needed to approximate the curve with linear segments.
 \vspace{.1cm}
 \item \textbf{Interaction Strength} Quantifying the impact of interaction effects is relevant when explanations are required, as most interpretability techniques use linear relationships to obtain explanations. Interaction Strength is measured as the fraction of 
 variance that can not be explained by main effects.

 \item \textbf{Sparsity} can be a desired property in case a simple explanation of a model is required, or obtaining features is costly and can potentially be avoided. In this work we measure sparsity as the fraction of features used.
\end{itemize}
A different approach towards achieving interpretability, would to instead focus on limiting an AutoML system to models, that are inherently interpretable. As those models rarely achieve optimal performances and trade-offs between interpretability and predictive perfromance cannot be assessed, we resort instead to look for models that are well-suited for post-hoc interpretability.

\paragraph{Fairness} has been established as a relevant criterion in Machine Learning when humans are subject to algorithmic decisions. The aim of the field is to encourage models that do not discriminate between certain sub-populations in the data.
\cite{hardt2016equality} define the concept of \textit{equalized odds} and \textit{equal opportunity}.
Given a protected attribute $A$ (e.g.  gender), an outcome $X$, a binary predictor $Y_b$, several criteria can be derived.
\begin{itemize} 
 \item \textbf{Independence}
   or equalized odds can be measured as follows:
    \[
    Pr\{Y_b = 1 | A = 0, Y = 1\} = Pr\{Y_b = 1 | A = 1, Y = 1\}
    \]
    i.e. if the true positive rate is equal in sub-populations indicated by $A$.
     \vspace{.1cm}
 \item \textbf{Sufficiency}
    or equality of opportunity can be measured as follows:
     \[
    Pr\{Y_b = 1 | A = 0, Y = y\} = Pr\{Y_b = 1 | A = 1, Y = y\}, y \in \{0, 1\}
    \]
    i.e. if the false positive and the false negative rates are equal in sub-populations.
     \vspace{.1cm}
 \item \textbf{Calibration} is another desirable criterion for classifier, especially in the context of fairness, where we might want to have calibrated probabilities in all groups. \cite{pleiss17} show, that models that are well-calibrated but also have equalized odds are only possible in case the predictor is \textit{perfect}, i.e does not make any errors.
\end{itemize}
\cite{vzliobaite2017measuring} provide a review of various discrimination measures that can be used in this context. 
A score for fairness can now be derived for example from the absolute differences of the given measure in each subgroup. In the use-case below, we use the differences in F1-Scores as a measure we want to minimize. The F1 score is the harmonic mean between the True Positive Rate and the Positive predictive value, and thus trades off true positives, false negatives and false positives.\\
 
\begin{center}

Addendum\footnote{\textcolor{red}{The paragraph was added after presentation at ECML-PKDD ADS 2019 Workshop. Nonetheless, the authors deem it important to make the statement in order to avoid harm caused by biased machine learning systems. An excellent resource for additional information is e.g. 2020 CVPR Tutorial on Fairness Accountability Transparency and Ethics in Computer Vision /\url{https://sites.google.com/view/fatecv-tutorial})}}\\

\fbox{\begin{minipage}{\textwidth}
 The use of machine learning in situations where individuals are affected by model decisions harbors opportunities as well as dangers. In the context of fairness, ML models can exhibit bias, similarly to humans. But in opposition to humans, bias in ML systems can often be explicitly measured. On the other hand, humans can be asked to justify or explain decisions, while this does not necessarily hold for ML models.
It is important to note that fairness can not be achieved solely through a reduction into mathematical criteria e.g. statistical and individual notions of fairness such as disparate treatment, disparate impact etc. . Many problems with such metrics still persist and require additional research. Furthermore, practitioners need not only take into account the model itself, but also the data used to train the algorithm, the process behind the collection and labeling of such data and eventual feedback loops arising from use of potentially biased models. It is of utmost importance to scrutinize data and resulting models from the perspective of all sub-groups (and intersections of those) in order to avoid introducing bias and causing harm to individuals.
\end{minipage}}
\end{center}

\paragraph{Robustness} as a concept, describes the behaviour of machine learning algorithms in situations where the data originally used to train a model is changed. A formal definition of robustness is currently lacking, which might arise from the many different concepts such a definition would need to cover. ~\cite{bousquet2002} define the notion of stability, which essentially measures how much a pre-defined loss-function deteriorates if an observation is held-out during training. This is not exactly what we are interested in as it requires extensive retraining. Instead we require a \textit{post-hoc} method that operates on a fitted model and training or testing data. A different approach, also coined \textit{stability} is provided in ~\cite{lange2003stability}. Their notion of stability measures the disagreement between a trained model on training data and test data.
In this work, we detail three measures of robustness, which we deem helpful in certain situations.
\begin{itemize} 
 \item \textbf{Perturbations} A very simple measure of robustness could be a classifiers' robustness to minimal perturbations in the input data.
 We create a copy $X^\star$ of our data $X$ by adding a small magnitude noise $N(0, \epsilon)$ scaled by $\epsilon$, typically $0.001 - 0.01$ times the range of the numerical feature. The robustness to perturbations can then be measured via the absolute difference of some loss $L$, for example \textit{accuracy}.\\
 \[
 |L(X,Y) - L(X^\star,Y)|
 \]
  \vspace{.1cm}
 \item \textbf{Adversarial Examples}
 A widely researched area of robustness is the field of Adversarial Examples ~\cite{szegedy2013intriguing, Papernot15}. Various different \textit{adversarial} attacks and defenses against such attacks have been proposed. 
 A variety of robustness measures can be derived from the different types of attacks proposed. 
 \vspace{.1cm}
 \item \textbf{Distribution shift} is a concept that is gathering widespread interest not only as a research field ~\cite{Zhang13}, but also as a problem in AutoML, which became evident from the AutoML Challenge organized at the NIPS 2018 conference ~\cite{guyon2019}. To the author's knowledge, no measure that serves as a proxy for a model's robustness to distribution shift is available.
\end{itemize}

\paragraph{Inference Time and Memory requirements} have been widely used as a measurement of the performance of machine learning algorithms. The time required for inference can be incorporated as a criterion. 

\paragraph{Sparsity} is also an important desideratum in other contexts, where interpretability is not necessarily required. In cases, where observing each feature incurs different costs, a user might want to find a model that achieves optimal performances using as few features as possible.

\begin{figure}[ht]
\centering
\includegraphics[width=\textwidth]{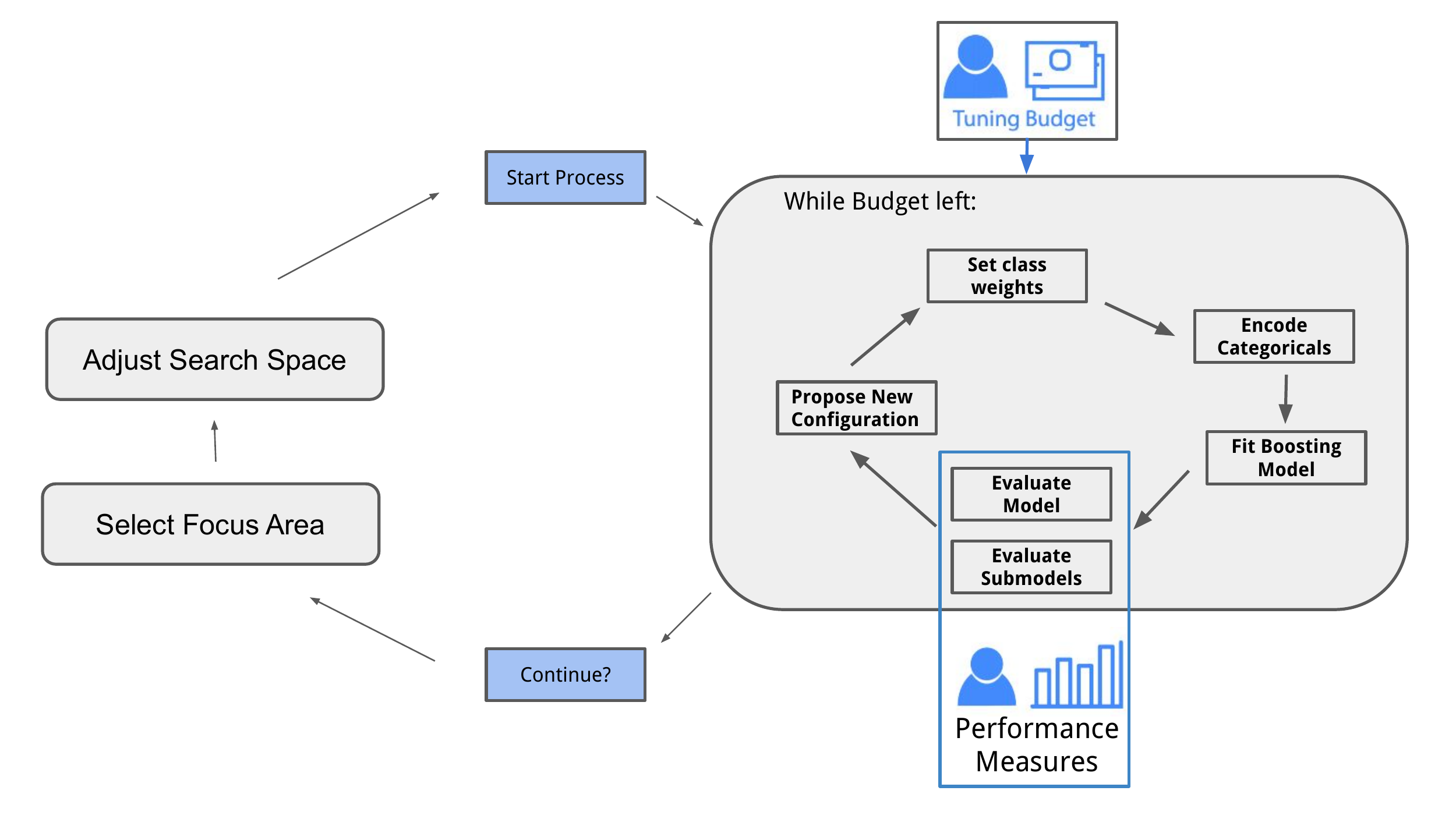}
\caption{Workflow for multi-criteria Autoxgboost. The user selects a measure and a budget, starts the process and then adapts the optimization before starting further evaluations.}
\centering
\label{fig:workflow}
\end{figure}

\section{Method}


This section introduces the structure of a first simple approach for multi-criteria AutoML. We heavily base our software on ~\cite{autoxgboost}, and include several design choices, such as the selection of preprocessing steps. The general workflow is detailed in Figure \ref{fig:workflow}. The implementation can be obtained from github \footnote{\url{https://github.com/pfistfl/autoxgboostMC}}.
Automatic gradient boosting simplifies AutoML to a fixed choice of machine learning algorithm by only using gradient boosting with trees (GBT).
Gradient Boosted Decision Trees are widely successfull for learning on tabular data and have desirable properties for AutoML systems, as they can deal with missing observations, are insensitive to outliers and can handle large amounts of features and data points. Additionally they are numerically stable and memory efficient. Additionally modern GBT frameworks like \textit{xgboost} ~\cite{xgboost} or \textit{lightgbm} ~\citep{lightgbm} are highly configurable with a large number of hyperparameters for regularization and optimization. As a result, they can approximate or cover many other scenarios, such as decision trees, random forests or linear models. Categorical feature transformation is performed as a preprocessing step.
We employ Sequential model-based optimization (SMBO), also known as Bayesian Optimization
as a hyperparameter optimization strategy ~\citep{Snoek2012}. The hyperparameter space we optimize is identical to ~\cite{autoxgboost}. We use multi-criteria Bayesian Optimization (~\cite{mlr, mlrMBO, horn2016multi}) (c.f section \ref{sec:mbo}) in order to optimize the machine-learning pipeline.

\subsection{Sub-evaluations}

In the context of multi-criteria optimization, early stopping is no longer trivial, as multiple pareto-optimal solutions might exist. The same holds for the selection of an optimal classification threshold as a postprocessing step in case a measure requiring binary outcomes instead of probabilities. At the same time, evaluating different thresholds or a sub-model using only a fraction of a model's gradient boosting iterations is very cheap after fitting a full model. In order to make use of this information we adopt the following procedure:

\begin{algorithm}
\label{algo:bayesopt}
\caption{Bayesian Optimization using sub-evaluations}
\begin{algorithmic}
\Require
    \State $S_m \gets \bigcup_1^m (\theta^\star_i, y_i)$: $m$ Initial evaluations
    \State $j \gets m + 1$
\While{Budget left}
    \State $\theta^\star_j \gets$ proposed using Bayesian Optimization on $S_{j-1}$
    \State $f_{\theta^\star,j} \gets$ fitted on data using $ \theta_j, nrounds_j$ and applying $thr_j$.
    \State $y_j \gets$ obtained by evaluating  $f_{\theta^\star,j}$ for each measure.
    \State $S_j \gets S_{j-1} \cup (\theta^\star_j, y_j)$
    \State $S_{sub, j} \gets$ obtain sub-evaluations (Algorithm $2$)
    \State $S_{sub, j} \gets$ keep only $S_{sub, j}$ which are on the pareto front of $S_j \cup S_{sub, j}$
    \State $S_j \gets S_j \cup S_{sub, j}$
    \State $j \gets j + 1$
\EndWhile
\end{algorithmic}
\end{algorithm}

A full pipeline configuration $\theta^\star \in \Theta^\star$ is composed of a threshold $thr \in [0;1]$, the number of boosting iterations $nrounds$ and several other pipeline hyperparameters, denoted by $\theta$ for simplicity.
From a set of $m$ randomly chosen initial configurations and their corresponding performances $S_m$ we start multi-criteria Bayesian Optimization as described in Algorithm $1$. The method for obtaining sub-evaluations is described in Algorithm $2$. 
In order to decrease the number of sub-evaluations, we resort to evaluating only $25\%, 50\%, 75\%$ and $90\%$ of $nrounds$ (rounded to the next integer).
Although this section describes the case of classifying a \textit{binary} target variable, extensions to multi-class classification can be made by instead using a vector
of thresholds instead.

\begin{algorithm}
\label{algo:subevals}
\caption{Obtaining Sub-evaluations}
\begin{algorithmic}
\Require
    \State Model $f_{\theta^\star,j}$; $i \gets 1$
\For{n in \{1, ..., nrounds\}}
\For{$thr$ in $\{0, 0.1, 0.2, ..., 1\}$}
    \State $S_i \gets$ evaluate $f_\theta^\star$ using $n$ iterations, applying threshold $thr$
    \State $i \gets i + 1$
\EndFor
\EndFor
\State $S_{sub} \gets \bigcup^i_1 S_i$
\end{algorithmic}
\end{algorithm}

\subsection{Multi-Criteria Bayesian Optimization}
\label{sec:mbo}
Formally multi-criteria optimization problems are defined by a set of target functions $f(\theta) = (f_1(\theta), \dots, f_k(\theta))$ which should be optimized simultaneously.
As there is no inherent order between the targets, the concept of \emph{Pareto dominance} is used to rank different candidate configurations.
One configuration $\theta$ pareto-dominates another configuration $\tilde\theta$, $\theta \preceq \tilde \theta$, if $f_i(\theta) \leq f_i(\tilde \theta)$ for $i = 1, \ldots, k$ and $\exists\, j\, f_j(\theta) < f_j(\tilde \theta)$, i.e., $\theta$ needs to be as good as $\tilde\theta$ in each component and strictly better in at least one.
A configuration $\theta$ is said to be \emph{non-dominated} if it is not dominated by any other configuration.
The set of all non-dominated points is the \emph{Pareto set}, which contains all trade-off solutions. 
Finally, the \emph{Pareto front} is evaluation of all configurations in the \emph{Pareto set}.
The goal of multi-criteria optimization is to learn the Pareto set.
There exists a plethora of different ways to extend Bayesian optimization to the multi-criteria case.
We choose \textbf{parEgo} ~\cite{parego}, as it is a simple method, and it naturally lends itself to focussing regions of the pareto front. \textbf{parEgo} is a rather simple extension, which scalarizes the set of target functions by using the augmented Tchebycheff norm 
\[
\max_{i=1,\dots,k}(w_if_i(\theta)) +\rho\sum_{i=i}^kw_if_i(\theta),
\]
with a different uniformly sampled weight vector $w$ such that $\sum_{i=1}^kw_i=1$ in each iteration. 
The augmentation term $\rho\sum_{i=i}^kw_if_i(\theta); \rho > 0$ is used to guarantee pareto-optimal solutions  \citep{miettinen2002scalarizing}.
This allows to apply standard single-criteria Bayesian optimization to the scalarized target function.
Furthermore the use of the augmented Tchebycheff norm allows to exclude regions of the pareto front which are not of practical relevance, e.g., models with extremely low predictive accuracy do not have to be considered regardless of their interpretability or fairness (\cite{Steuer},~\cite{Hakanen2017OnUD}). 
This is done by constraining the values of some $w_i$ between certain values.
We adapt a similar procedure, where the user can choose ranges for the weights $w_i$, such that the algorithm focuses on a selected region of the pareto front (see e.g. the blue line in Figure \ref{fig:initial_iters}).\\
For the case of $k = 2$ objectives, the weight vector $w \in [0;1]^k$ can range from $(1, 0)$ (only optimize first objective) to $(0, 1)$ (only optimizing the second objective). By limiting $w$ to $[l, 1-l] \times [u, 1-u]; 0 < l < u < 1$ we can effectively limit the possible trade-offs we might be willing to make.

\vspace{-.2cm}
\section{Application: A Fair Model for Income Prediction}
Income prediction of employers is a versatile use-case for the application of multi-criteria AutoML, as several important criteria for a model can be derived.
While trying to minimize the missclassification error, moral and ethical principles must also be adhered to. 
Thus, a model cannot be biased or unfair towards different sub-populations, e.g. men can not be systematically favoured over women regarding their income.
As a third possible criterion, we might require an interpretable model, as a model might need to be accepted by regulatory bodies.\\

\vspace{-4pt}
\begin{figure}[t!]
\begin{subfigure}[t]{0.50\textwidth}
\includegraphics[width=\linewidth]{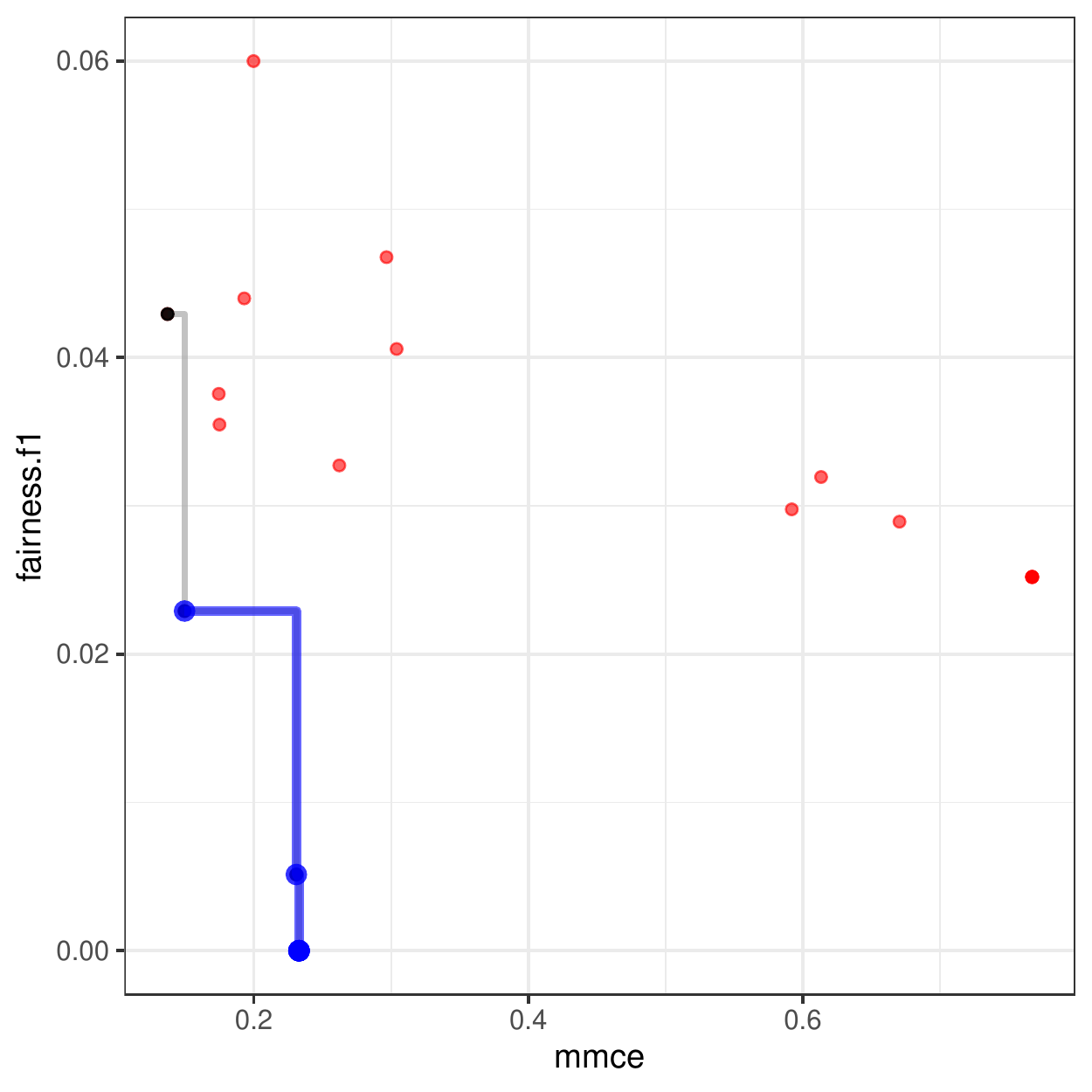}
\subcaption{Full pareto front of first AutoxgboostMC run. Not pareto-optimal points are displayed in red.}
\label{fig:initial_iters}
\end{subfigure}
\begin{subfigure}[t]{0.50\textwidth}
\includegraphics[width=\linewidth]{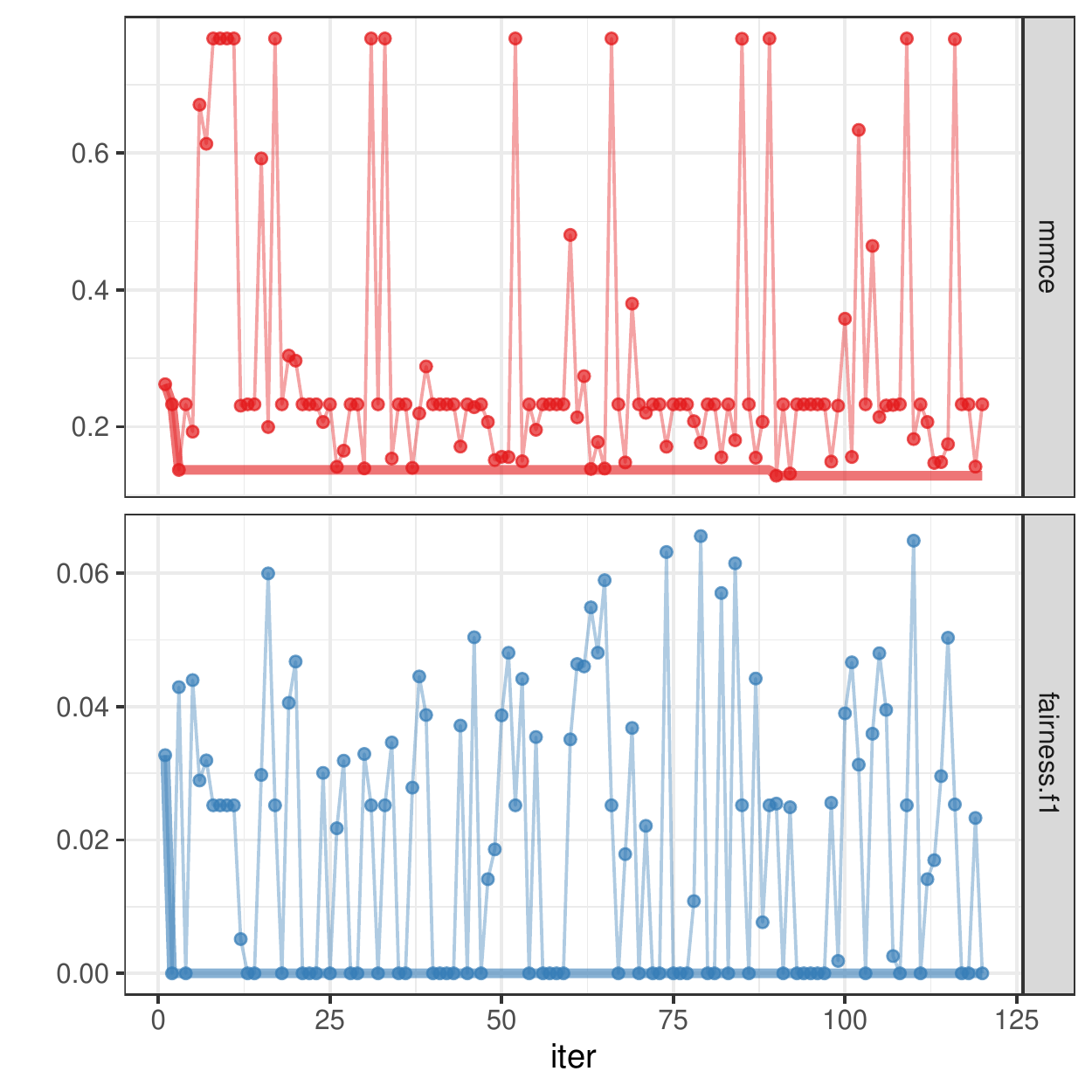}
\subcaption{Final optimization path.}
\label{fig:opt_path}
\end{subfigure}
\end{figure}

We use a fairness measure described in section \ref{sec:measures}, namely the absolute difference in F1-Scores between two sub-populations \textit{male} and \textit{female}. In order to start the AutoML process, we simply need to specify our tuning budget by either setting the number of MBO-iterations or the desired time for tuning. To get a first impression of the pareto front, we start with a tuning budget of only $20$ iterations. Figure~\ref{fig:initial_iters} shows the resulting pareto front. We can now use this, to focus the search towards trade-offs we are interested in. This is done by limiting the range of projections available to \textbf{parEgo}. For a first investigation, we choose values between $0.1$ and $0.9$. Those lower and upper limits on the projections can be adapted throughout the process.

\begin{figure}[b!]
\centering
\includegraphics[width=\linewidth]{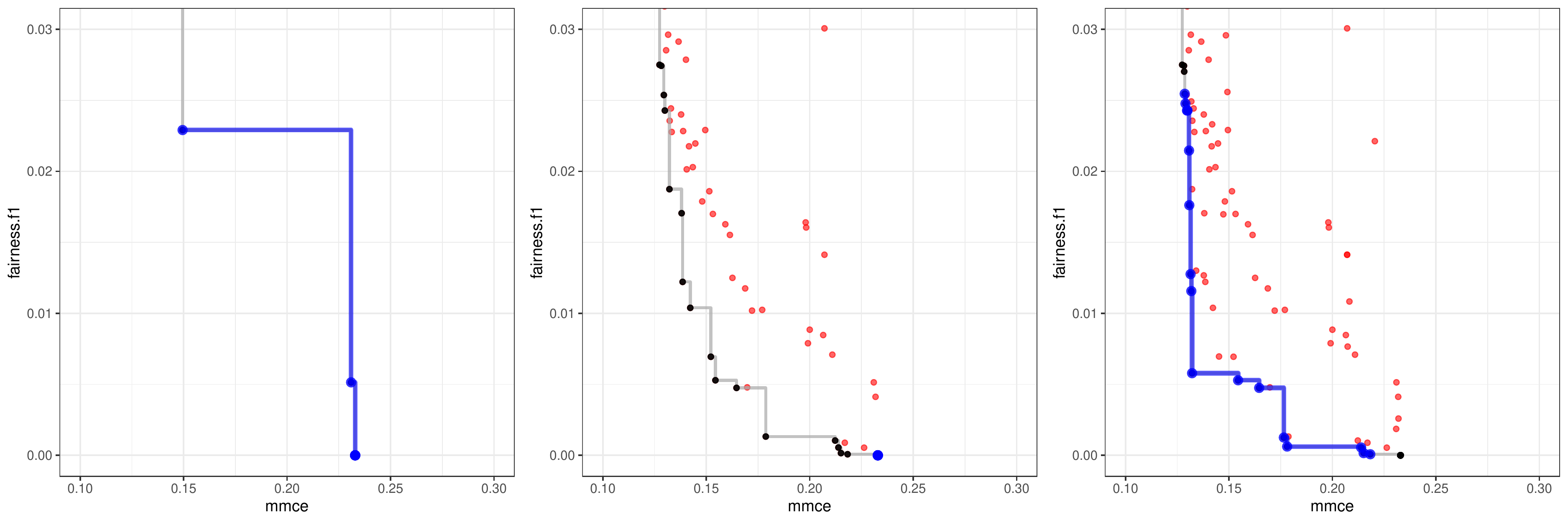}
\caption{Pareto fronts after each AutoxgboosMC run ($20$, $70$, $120$ tuning iterations). Zoomed in, in order to better show the pareto-front. The region we focus on is coloured in blue.}
\centering
\label{fig:pareto_plots}
\end{figure} 

\begingroup
\begin{wraptable}{rb}{6.1cm}
\centering
\begin{tabular}{|l|rr|rr|}
\toprule
       &      Valid. &           &  Test & \\
Method & mmce & fairF1 & mmce & fairF1\\
\hline
autoxgboost   & \textbf{0.125}     &     -      & 0.165  & 0.038  \\
PF (ours) & 0.129 & 0.023 & 0.139 & 0.061\\

PF (ours) & 0.129 & 0.020 & 0.131 & 0.064\\

PF (ours) & 0.130 & 0.002 & \textbf{0.130} & 0.080\\

PF (ours) & 0.131 & 0.001 & 0.159 & 0.059\\

PF (ours) & 0.132 & 0.001 & 0.157 & 0.067\\

PF (ours) & 0.133 & 0.001 & 0.156 & 0.060\\

PF (ours) & 0.135 & \textbf{0.000} & 0.142 & 0.096\\

PF (ours) & 0.137 & \textbf{0.000} & 0.147 & 0.059\\

PF (ours) & 0.142 & \textbf{0.000} & 0.146 & 0.077\\

PF (ours) & 0.158 & \textbf{0.000} & 0.284 & 0.116\\

PF (ours) & 0.238 & \textbf{0.000} & 0.244 & \textbf{0.000}\\
\bottomrule
\end{tabular}
\caption{Performances of models from the pareto-front on held-out test set. We compare solutions from the Pareto front (PF) to \cite{autoxgboost} (optimizing mmce) after 120 iterations.}
\label{tab:pareto_front}
\end{wraptable}

Afterwards, we can simply continue training with additional budget.
Continuing this training twice, we can also access the final optimization path, which is shown in Figure~\ref{fig:opt_path}.
For each chosen measure, it shows the achieved performance for each function evaluation, and the (single-criteria) optimum achieved.
Finally, we observe the pareto fronts as illustrated in Figure~\ref{fig:pareto_plots} for final tuning after $20$, $70$ and $120$ iterations.\\
We compare to \cite{autoxgboost}, optimizing a single objective (mmce). Note that solely optimizing for Fairness is not sensible, as many models achieve a fairness score of 0 (on validation data). 
The user can then choose an optimal hyperparameter configuration from the pareto front which matches her preferences best.
Table \ref{tab:pareto_front} displays different points from the pareto front as well as the single-objective method evaluated on test data.

\endgroup

\section{Outlook}

In this work we conduct a first investigation into AutoML systems that can optimize a machine learning pipeline with respect to many different criteria. We provide several measures, that can be used as proxies for concepts such as \textit{Fairness, Interpretability, Robustness} and others. Additionally, we implement a simplified AutoML system, that can optimize multiple objectives simultaneously, and can therefore serve as a tool for investigating such scenarios.
The potential and necessity of our approach is demonstrated in a use-case. 

The proposed method can be extended and improved in multiple directions.
In a first iteration, we aim to include a wider array of gradient boosting methods, such as LightGBM (\cite{lightgbm}) and catboost (\cite{catboost}) into our framework. By combining this with a larger set of different pre- and post-processing methods, which can be tailored towards improving the different measures listed in section \ref{sec:measures}, we hope to obtain a toolbox that is suitable for many different situations where multiple criteria are required. 
Several interesting enhancements to the optimization procedure could also be made, either by adopting promising approaches from Bayesian Optimization (c.f. \cite{multicritMBO}), or by adopting other search procedures.

The real underlying preferences a user has towards selecting a model might not always be easily quantify-able, because they rely on previous experience, implementation or other details. At the same time, a user can be asked to provide (\textit{noisy}) labels for a set of models or to indicate preferences of one model over another (c.f \cite{gonzalez2017preferential}). 
In future research, this might serve as an interesting avenue towards more human-centered AutoML.
A third important part of research we aim to conduct is towards making AutoML methods more readily available to other user-groups, while at the same time providing them with sufficient tools to obtain models tailored towards the specific applications needs. In order to achieve this, we aim to research User Interfaces that make the AutoML more transparent to the user, while at the same time ensuring reproducibility.

~\\\noindent{}\textbf{Acknowledgements.}
This work has been funded by the German Federal Ministry of Education and Research (BMBF) under Grant No. 01IS18036A. 

\bibliography{refs}

\begin{thebibliography}{}

\bibitem[\protect\astroncite{Apley}{2016}]{apley2016visualizing}
Apley, D.~W. (2016).
\newblock Visualizing the effects of predictor variables in black box
  supervised learning models.
\newblock {\em arXiv preprint arXiv:1612.08468}.

\bibitem[\protect\astroncite{Barocas et~al.}{2018}]{fairmlbook}
Barocas, S., Hardt, M., and Narayanan, A. (2018).
\newblock {\em Fairness and Machine Learning}.
\newblock fairmlbook.org.
\newblock \url{http://www.fairmlbook.org}.

\bibitem[\protect\astroncite{Bischl et~al.}{2016}]{mlr}
Bischl, B., Lang, M., Kotthoff, L., Schiffner, J., Richter, J., Studerus, E.,
  Casalicchio, G., and Jones, Z.~M. (2016).
\newblock {mlr}: Machine learning in {R}.
\newblock {\em Journal of Machine Learning Research}, 17(170):1--5.

\bibitem[\protect\astroncite{Bischl et~al.}{2018}]{mlrMBO}
Bischl, B., Richter, J., Bossek, J., Horn, D., Thomas, J., and Lang, M. (2018).
\newblock {{mlrMBO}}: {{A Modular Framework}} for {{Model}}-{{Based
  Optimization}} of {{Expensive Black}}-{{Box Functions}}.

\bibitem[\protect\astroncite{Blot et~al.}{2016}]{blot16}
Blot, A., Hoos, H.~H., Vermeulen-Jourdan, L., Kessaci-Marmion, M.-{\'E}., and
  Trautmann, H. (2016).
\newblock Mo-paramils: A multi-objective automatic algorithm configuration
  framework.
\newblock In {\em LION}.

\bibitem[\protect\astroncite{Bousquet and Elisseeff}{2002}]{bousquet2002}
Bousquet, O. and Elisseeff, A. (2002).
\newblock Stability and generalization.
\newblock {\em J. Mach. Learn. Res.}, 2:499--526.

\bibitem[\protect\astroncite{Chen and Guestrin}{2016}]{xgboost}
Chen, T. and Guestrin, C. (2016).
\newblock {XGBoost}: A scalable tree boosting system.
\newblock In {\em Proceedings of the 22nd ACM SIGKDD International Conference
  on Knowledge Discovery and Data Mining}, KDD '16, pages 785--794, New York,
  NY, USA. ACM.

\bibitem[\protect\astroncite{Dorogush et~al.}{2017}]{catboost}
Dorogush, A.~V., Ershov, V., and Gulin, A. (2017).
\newblock Catboost: gradient boosting with categorical features support.

\bibitem[\protect\astroncite{Everson and Fieldsend}{2006}]{everson2006multi}
Everson, R.~M. and Fieldsend, J.~E. (2006).
\newblock Multi-class roc analysis from a multi-objective optimisation
  perspective.
\newblock {\em Pattern Recognition Letters}, 27(8):918--927.

\bibitem[\protect\astroncite{Fan et~al.}{2017}]{fan2017comparative}
Fan, Z., Fang, Y., Li, W., Lu, J., Cai, X., and Wei, C. (2017).
\newblock A comparative study of constrained multi-objective evolutionary
  algorithms on constrained multi-objective optimization problems.
\newblock In {\em 2017 IEEE Congress on Evolutionary Computation (CEC)}, pages
  209--216. IEEE.

\bibitem[\protect\astroncite{Feurer et~al.}{2015}]{autosklearn}
Feurer, M., Klein, A., Eggensperger, K., Springenberg, J., Blum, M., and
  Hutter, F. (2015).
\newblock Efficient and robust automated machine learning.
\newblock In Cortes, C., Lawrence, N.~D., Lee, D.~D., Sugiyama, M., and
  Garnett, R., editors, {\em Advances in Neural Information Processing Systems
  28}, pages 2962--2970. Curran Associates, Inc.

\bibitem[\protect\astroncite{Friedman}{2001}]{gradientboosting}
Friedman, J.~H. (2001).
\newblock Greedy function approximation: A gradient boosting machine.
\newblock {\em Ann. Statist.}, 29(5):1189--1232.

\bibitem[\protect\astroncite{Gonz{\'a}lez
  et~al.}{2017}]{gonzalez2017preferential}
Gonz{\'a}lez, J., Dai, Z., Damianou, A., and Lawrence, N.~D. (2017).
\newblock Preferential bayesian optimization.
\newblock In {\em Proceedings of the 34th International Conference on Machine
  Learning-Volume 70}, pages 1282--1291. JMLR. org.

\bibitem[\protect\astroncite{Guyon et~al.}{2019}]{guyon2019}
Guyon, I., Sun-Hosoya, L., Boull{\'e}, M., Escalante, H.~J., Escalera, S., Liu,
  Z., Jajetic, D., Ray, B., Saeed, M., Sebag, M., Statnikov, A., Tu, W.-W., and
  Viegas, E. (2019).
\newblock {\em Analysis of the AutoML Challenge Series 2015--2018}, pages
  177--219.
\newblock Springer International Publishing, Cham.

\bibitem[\protect\astroncite{Hakanen and Knowles}{2017}]{Hakanen2017OnUD}
Hakanen, J. and Knowles, J.~D. (2017).
\newblock On using decision maker preferences with parego.
\newblock In {\em EMO}.

\bibitem[\protect\astroncite{Hall et~al.}{2009}]{weka}
Hall, M., Frank, E., Holmes, G., Pfahringer, B., Reutemann, P., and Witten,
  I.~H. (2009).
\newblock The weka data mining software: An update.
\newblock {\em SIGKDD Explor. Newsl.}, 11(1):10--18.

\bibitem[\protect\astroncite{Handl et~al.}{2007}]{handl2007multiobjective}
Handl, J., Kell, D.~B., and Knowles, J. (2007).
\newblock Multiobjective optimization in bioinformatics and computational
  biology.
\newblock {\em IEEE/ACM Transactions on Computational Biology and
  Bioinformatics (TCBB)}, 4(2):279--292.

\bibitem[\protect\astroncite{Hardt et~al.}{2016}]{hardt2016equality}
Hardt, M., Price, E., Srebro, N., et~al. (2016).
\newblock Equality of opportunity in supervised learning.
\newblock In {\em Advances in neural information processing systems}, pages
  3315--3323.

\bibitem[\protect\astroncite{Hern{\'a}ndez-Lobato
  et~al.}{2016}]{hernandez2016general}
Hern{\'a}ndez-Lobato, J.~M., Gelbart, M.~A., Adams, R.~P., Hoffman, M.~W., and
  Ghahramani, Z. (2016).
\newblock A general framework for constrained bayesian optimization using
  information-based search.
\newblock {\em The Journal of Machine Learning Research}, 17(1):5549--5601.

\bibitem[\protect\astroncite{Horn and Bischl}{2016}]{horn2016multi}
Horn, D. and Bischl, B. (2016).
\newblock Multi-objective parameter configuration of machine learning
  algorithms using model-based optimization.
\newblock In {\em Computational Intelligence (SSCI), 2016 IEEE Symposium Series
  on}, pages 1--8. IEEE.

\bibitem[\protect\astroncite{Howard et~al.}{2017}]{MobileNet}
Howard, A.~G., Zhu, M., Chen, B., Kalenichenko, D., Wang, W., Weyand, T.,
  Andreetto, M., and Adam, H. (2017).
\newblock Mobilenets: Efficient convolutional neural networks for mobile vision
  applications.
\newblock {\em CoRR}, abs/1704.04861.

\bibitem[\protect\astroncite{Huang et~al.}{2016}]{SpeedAccObjDetect}
Huang, J., Rathod, V., Sun, C., Zhu, M., Korattikara, A., Fathi, A., Fischer,
  I., Wojna, Z., Song, Y., Guadarrama, S., and Murphy, K. (2016).
\newblock Speed/accuracy trade-offs for modern convolutional object detectors.
\newblock {\em CoRR}, abs/1611.10012.

\bibitem[\protect\astroncite{Hutter et~al.}{2011}]{smac}
Hutter, F., Hoos, H.~H., and Leyton-Brown, K. (2011).
\newblock {\em Sequential Model-Based Optimization for General Algorithm
  Configuration}, pages 507--523.
\newblock Springer Berlin Heidelberg, Berlin, Heidelberg.

\bibitem[\protect\astroncite{Jin and Sendhoff}{2008}]{jin2008pareto}
Jin, Y. and Sendhoff, B. (2008).
\newblock Pareto-based multiobjective machine learning: An overview and case
  studies.
\newblock {\em IEEE Transactions on Systems, Man, and Cybernetics, Part C
  (Applications and Reviews)}, 38(3):397--415.

\bibitem[\protect\astroncite{Johnson et~al.}{2017}]{FAISS}
Johnson, J., Douze, M., and J{\'{e}}gou, H. (2017).
\newblock Billion-scale similarity search with gpus.
\newblock {\em CoRR}, abs/1702.08734.

\bibitem[\protect\astroncite{Ke et~al.}{2017}]{lightgbm}
Ke, G., Meng, Q., Finley, T., Wang, T., Chen, W., Ma, W., Ye, Q., and Liu,
  T.-Y. (2017).
\newblock Lightgbm: A highly efficient gradient boosting decision tree.
\newblock In Guyon, I., Luxburg, U.~V., Bengio, S., Wallach, H., Fergus, R.,
  Vishwanathan, S., and Garnett, R., editors, {\em Advances in Neural
  Information Processing Systems 30}, pages 3149--3157. Curran Associates, Inc.

\bibitem[\protect\astroncite{Knowles}{2004}]{parego}
Knowles, J. (2004).
\newblock Parego: A hybrid algorithm with on-line landscape approximation for
  expensive multiobjective optimization problems.
\newblock Technical Report TR-COMPSYSBIO-2004-01, University of Manchester.

\bibitem[\protect\astroncite{Lange et~al.}{2003}]{lange2003stability}
Lange, T., Braun, M.~L., Roth, V., and Buhmann, J.~M. (2003).
\newblock Stability-based model selection.
\newblock In {\em Advances in neural information processing systems}, pages
  633--642.

\bibitem[\protect\astroncite{Miettinen and
  M{\"a}kel{\"a}}{2002}]{miettinen2002scalarizing}
Miettinen, K. and M{\"a}kel{\"a}, M.~M. (2002).
\newblock On scalarizing functions in multiobjective optimization.
\newblock {\em OR spectrum}, 24(2):193--213.

\bibitem[\protect\astroncite{Molnar}{2019}]{molnar2019}
Molnar, C. (2019).
\newblock {\em Interpretable Machine Learning}.
\newblock \url{https://christophm.github.io/interpretable-ml-book/}.

\bibitem[\protect\astroncite{Molnar et~al.}{2019}]{molnar2019quantifying}
Molnar, C., Casalicchio, G., and Bischl, B. (2019).
\newblock Quantifying interpretability of arbitrary machine learning models
  through functional decomposition.
\newblock {\em arXiv preprint arXiv:1904.03867}.

\bibitem[\protect\astroncite{Olson et~al.}{2016}]{tpot}
Olson, R.~S., Urbanowicz, R.~J., Andrews, P.~C., Lavender, N.~A., Kidd, L.~C.,
  and Moore, J.~H. (2016).
\newblock Automating biomedical data science through tree-based pipeline
  optimization.
\newblock In Squillero, G. and Burelli, P., editors, {\em Applications of
  Evolutionary Computation}, pages 123--137, Cham. Springer International
  Publishing.

\bibitem[\protect\astroncite{Papernot et~al.}{2015}]{Papernot15}
Papernot, N., McDaniel, P.~D., Wu, X., Jha, S., and Swami, A. (2015).
\newblock Distillation as a defense to adversarial perturbations against deep
  neural networks.
\newblock {\em CoRR}, abs/1511.04508.

\bibitem[\protect\astroncite{Paria et~al.}{2018}]{multicritMBO}
Paria, B., Kandasamy, K., and P{\'{o}}czos, B. (2018).
\newblock A flexible multi-objective bayesian optimization approach using
  random scalarizations.
\newblock {\em CoRR}, abs/1805.12168.

\bibitem[\protect\astroncite{Pleiss et~al.}{2017}]{pleiss17}
Pleiss, G., Raghavan, M., Wu, F., Kleinberg, J., and Weinberger, K.~Q. (2017).
\newblock On fairness and calibration.
\newblock In {\em Proceedings of the 31st International Conference on Neural
  Information Processing Systems}, NIPS'17, pages 5684--5693, USA. Curran
  Associates Inc.

\bibitem[\protect\astroncite{Snoek et~al.}{2012}]{Snoek2012}
Snoek, J., Larochelle, H., and Adams, R.~P. (2012).
\newblock Practical bayesian optimization of machine learning algorithms.
\newblock In {\em Proceedings of the 25th International Conference on Neural
  Information Processing Systems - Volume 2}, NIPS'12, pages 2951--2959, USA.
  Curran Associates Inc.

\bibitem[\protect\astroncite{Steuer and Choo}{1983}]{Steuer}
Steuer, R.~E. and Choo, E.-U. (1983).
\newblock An interactive weighted tchebycheff procedure for multiple objective
  programming.
\newblock {\em Math. Program.}, 26(3):326--344.

\bibitem[\protect\astroncite{Szegedy et~al.}{2013}]{szegedy2013intriguing}
Szegedy, C., Zaremba, W., Sutskever, I., Bruna, J., Erhan, D., Goodfellow, I.,
  and Fergus, R. (2013).
\newblock Intriguing properties of neural networks.
\newblock {\em arXiv preprint arXiv:1312.6199}.

\bibitem[\protect\astroncite{Thomas et~al.}{2018}]{autoxgboost}
Thomas, J., Coors, S., and Bischl, B. (2018).
\newblock Automatic gradient boosting.
\newblock In {\em International Workshop on Automatic Machine Learning at
  ICML}.

\bibitem[\protect\astroncite{Thornton et~al.}{2013}]{autoweka}
Thornton, C., Hutter, F., Hoos, H.~H., and Leyton-Brown, K. (2013).
\newblock Auto-{WEKA}: Combined selection and hyperparameter optimization of
  classification algorithms.
\newblock In {\em Proc.~of KDD-2013}, pages 847--855.

\bibitem[\protect\astroncite{Wang et~al.}{2019}]{ATMSeer}
Wang, Q., Ming, Y., Jin, Z., Shen, Q., Liu, D., Smith, M.~J., Veeramachaneni,
  K., and Qu, H. (2019).
\newblock Atmseer: Increasing transparency and controllability in automated
  machine learning.
\newblock In {\em Proceedings of the 2019 CHI Conference on Human Factors in
  Computing Systems}, CHI '19, pages 681:1--681:12, New York, NY, USA. ACM.

\bibitem[\protect\astroncite{Wilson et~al.}{2018}]{AutoGBT}
Wilson, J., Meher, A.~K., Bindu, B.~V., Sharma, M., Pareek, V., Chaudhury, S.,
  and Lall, B. (2018).
\newblock Autogbt:automatically optimized gradient boosting trees for
  classifying large volume high cardinality data streams under concept-drift.
\newblock \url{ https://github.com/flytxtds/AutoGBT }.

\bibitem[\protect\astroncite{Zhang et~al.}{2013}]{Zhang13}
Zhang, K., Sch\"{o}lkopf, B., Muandet, K., and Wang, Z. (2013).
\newblock Domain adaptation under target and conditional shift.
\newblock In {\em Proceedings of the 30th International Conference on
  International Conference on Machine Learning - Volume 28}, ICML'13, pages
  III--819--III--827. JMLR.org.

\bibitem[\protect\astroncite{Zhang et~al.}{2015}]{sprint}
Zhang, T., Georgiopoulos, M., and Anagnostopoulos, G.~C. (2015).
\newblock Sprint multi-objective model racing.
\newblock In {\em Proceedings of the 2015 Annual Conference on Genetic and
  Evolutionary Computation}, GECCO '15, pages 1383--1390, New York, NY, USA.
  ACM.

\bibitem[\protect\astroncite{{\v{Z}}liobait{\.e}}{2017}]{vzliobaite2017measuring}
{\v{Z}}liobait{\.e}, I. (2017).
\newblock Measuring discrimination in algorithmic decision making.
\newblock {\em Data Mining and Knowledge Discovery}, 31(4):1060--1089.

\end{thebibliography}
\bibliographystyle{apa}

\clearpage
\end{document}